\documentclass[letterpaper, 10 pt, conference]{ieeeconf}  

\IEEEoverridecommandlockouts                              

\usepackage{graphicx}
\usepackage{amsmath,amssymb} 
\usepackage{xcolor}

\definecolor{rgb0}{RGB}{0,0,0}
\definecolor{rgb1}{RGB}{0,0,255}
\definecolor{rgb2}{RGB}{0,255,0}
\definecolor{rgb3}{RGB}{0,255,255}
\definecolor{rgb4}{RGB}{255,0,0}
\definecolor{rgb5}{RGB}{255,0,255}
\definecolor{rgb6}{RGB}{255,255,0}
\definecolor{rgb7}{RGB}{255,255,255}
\definecolor{rgb8}{RGB}{0,0,128}
\definecolor{rgb9}{RGB}{0,128,0}
\definecolor{rgb10}{RGB}{0,128,128}
\definecolor{rgb11}{RGB}{128,0,0}
\definecolor{rgb12}{RGB}{128,0,128}

\definecolor{rgb13}{RGB}{128,128,0}
\definecolor{rgb14}{RGB}{128,128,128}
\definecolor{rgb15}{RGB}{255,96,0}
\definecolor{rgb16}{RGB}{255,215,180}
\definecolor{rgb17}{RGB}{170,110,40}
\definecolor{rgb18}{RGB}{230,190,255}
\definecolor{rgb19}{RGB}{210,245,60}
\definecolor{rgb20}{RGB}{170,255,195}
\definecolor{rgb21}{RGB}{230,25,75}
\definecolor{rgb22}{RGB}{255,250,200}
\definecolor{rgb23}{RGB}{0,130,200}
\definecolor{rgb24}{RGB}{60,180,75}
\definecolor{rgb25}{RGB}{245,130,48}

\definecolor{rgb26}{RGB}{205,205,0}
\definecolor{rgb27}{RGB}{255,193,37}
\definecolor{rgb28}{RGB}{255,0,128}
\definecolor{rgb29}{RGB}{113,113,198}
\definecolor{rgb30}{RGB}{202,225,255}
\definecolor{rgb31}{RGB}{95,158,160}
\definecolor{rgb32}{RGB}{0,250,154}
\definecolor{rgb33}{RGB}{205,186,150}
\definecolor{rgb34}{RGB}{139,69,0}
\definecolor{rgb35}{RGB}{255,128,96}
\definecolor{rgb36}{RGB}{142,56,142}
\definecolor{rgb37}{RGB}{198,113,133}
\definecolor{rgb38}{RGB}{197,193,170}

\usepackage{url}

\usepackage{acronym}
\acrodef{PBR}{Physically Based Renderer}
\acrodef{PoV}{Point of View}
\acrodef{HRI}{Human-Robot Interaction}
\acrodef{FABRIK}{Forward And Backward Reaching Inverse Kinematics}
\acrodef{FPS}{Frames per Second}
\acrodef{MSAA}{Multisample Anti-Aliasing}
\acrodef{UE4}{Unreal Engine 4}
\acrodef{VR}{Virtual Reality}

\title{\LARGE \bf The RobotriX: An eXtremely Photorealistic and Very-Large-Scale Indoor Dataset of Sequences with Robot Trajectories and Interactions} 

\author{Alberto Garcia-Garcia, Pablo Martinez-Gonzalez, Sergiu Oprea,\\ John Alejandro Castro-Vargas, Sergio Orts-Escolano, Jose Garcia-Rodriguez and Alvaro Jover-Alvarez
\thanks{Albert, Pablo, Sergiu, John, Sergio, Jose and Alvaro are with 3D Perception Lab, University of Alicante, Spain
        {\tt\small agarcia@dtic.ua.es, pmartinez@dtic.ua.es, soprea@dtic.ua.es, jacastro@dtic.ua.es, sorts@ua.es, jgarcia@dtic.ua.es, ajover@dtic.ua.es}}%
}

\begin{document}

\maketitle

\thispagestyle{empty}
\pagestyle{empty}


\begin{abstract}

Enter the RobotriX, an extremely photorealistic indoor dataset designed to enable the application of deep learning techniques to a wide variety of robotic vision problems. The RobotriX consists of hyperrealistic indoor scenes which are explored by robot agents which also interact with objects in a visually realistic manner in that simulated world. Photorealistic scenes and robots are rendered by Unreal Engine into a virtual reality headset which captures gaze so that a human operator can move the robot and use controllers for the robotic hands; scene information is dumped on a per-frame basis so that it can be reproduced offline to generate raw data and ground truth labels. By taking this approach, we were able to generate a dataset of $38$ semantic classes totaling $8M$ stills recorded at $+60$ frames per second with full HD resolution. For each frame, RGB-D and 3D information is provided with full annotations in both spaces. Thanks to the high quality and quantity of both raw information and annotations, the RobotriX will serve as a new milestone for investigating 2D and 3D robotic vision tasks with large-scale data-driven techniques.

\end{abstract}


\begin{figure}[!htb]
  \centering
  \includegraphics[width=0.325\linewidth]{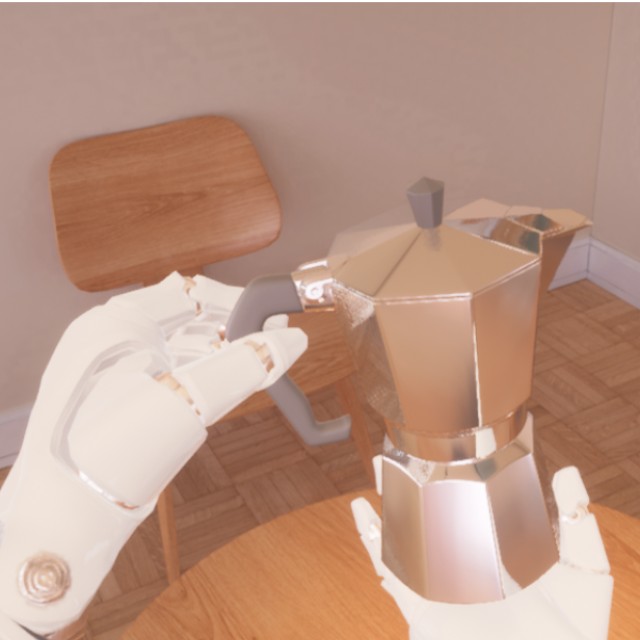}
  \includegraphics[width=0.325\linewidth]{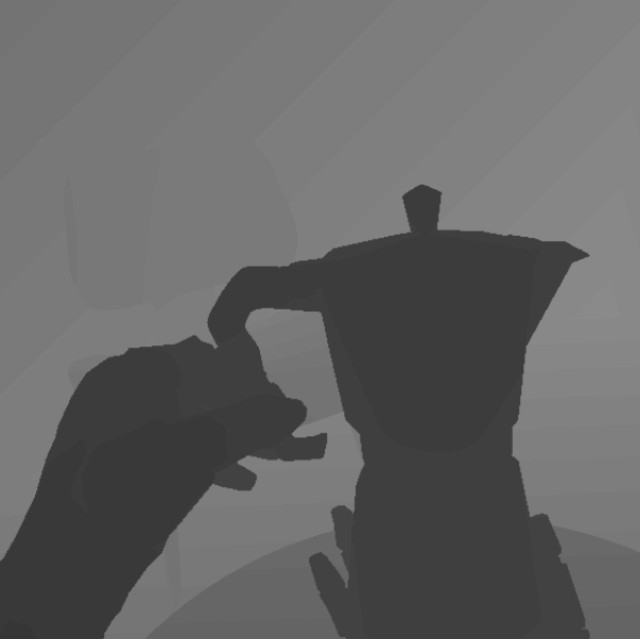}
  \includegraphics[width=0.325\linewidth]{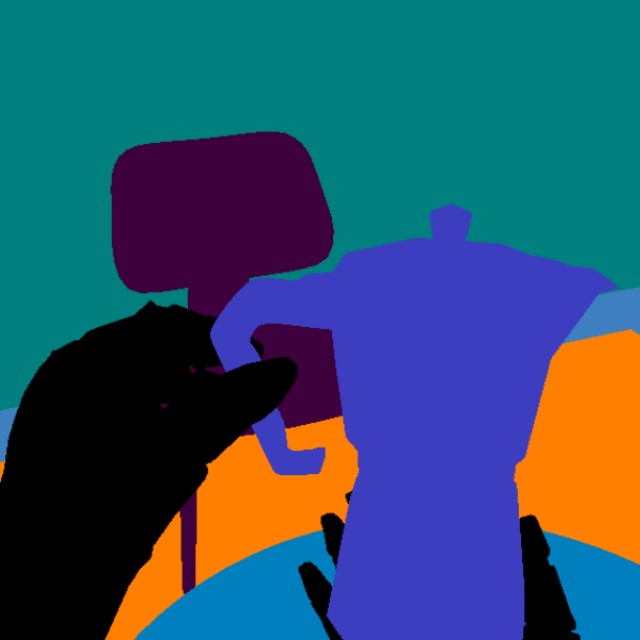}\\
  \smallskip
  \includegraphics[width=0.325\linewidth]{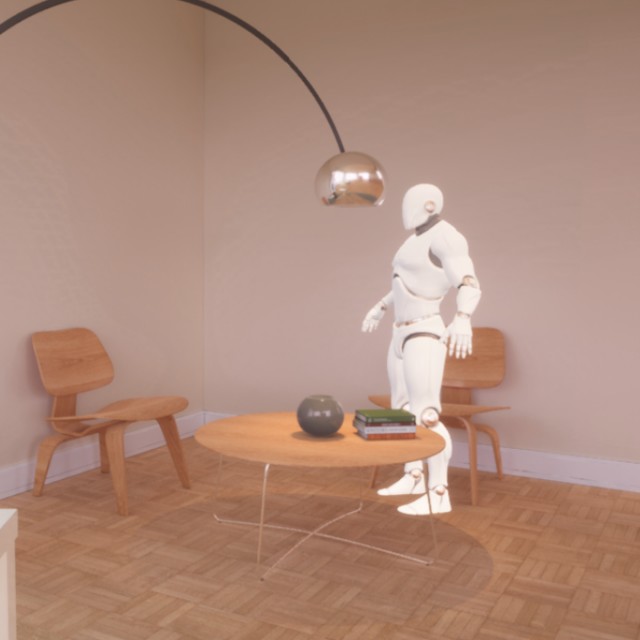}
  \includegraphics[width=0.325\linewidth]{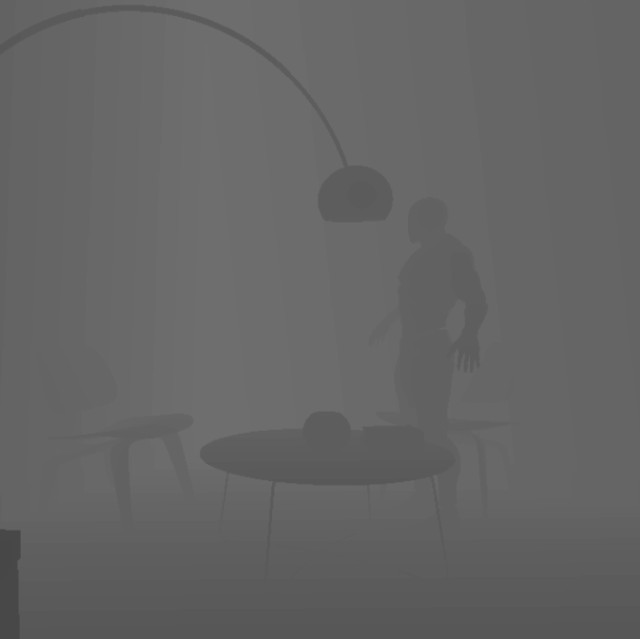}
  \includegraphics[width=0.325\linewidth]{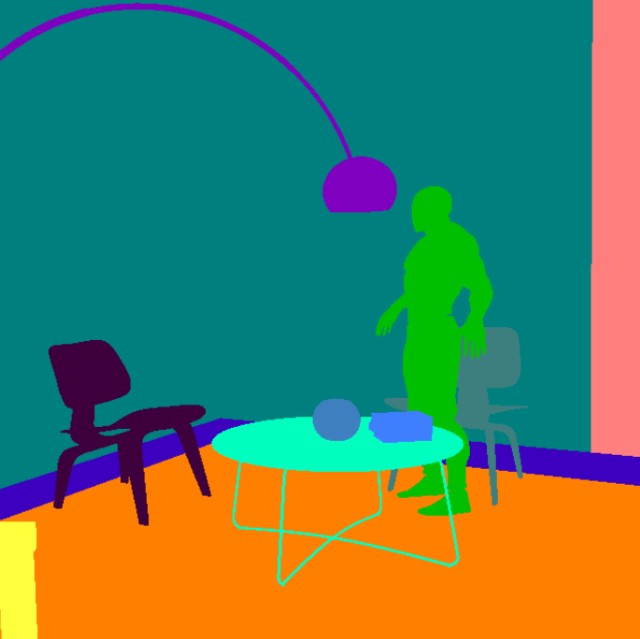}\\
  \smallskip
  \includegraphics[width=0.325\linewidth]{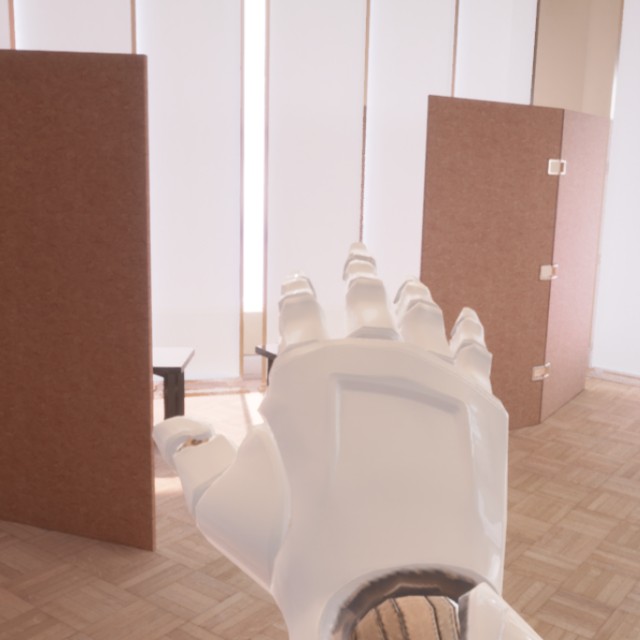}
  \includegraphics[width=0.325\linewidth]{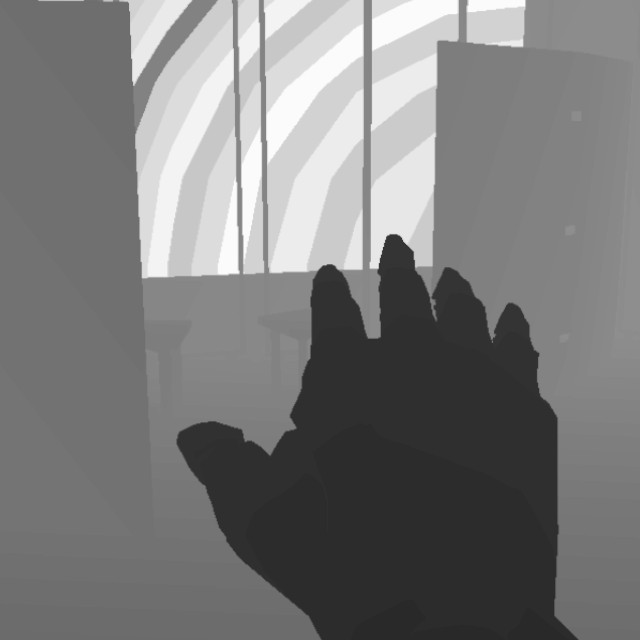}
  \includegraphics[width=0.325\linewidth]{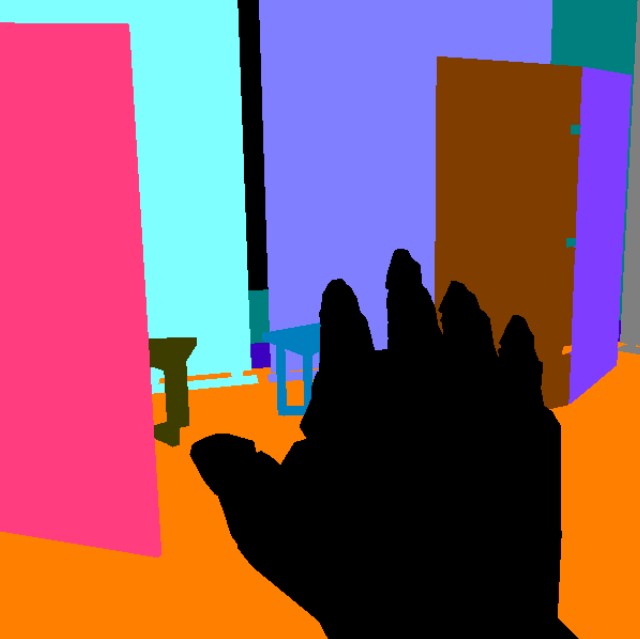}\\
  \caption{The RobotriX features extremely photorealistic indoor environments in which robot movements and interactions with objects are captured from multiple points of view at high frame rates and resolutions.}
  \label{fig:overview}
\end{figure}

\section{INTRODUCTION}
\label{sec:introduction}

Recent years have witnessed an increasing dominance of deep learning techniques targeted at a wide range of robotic vision problems such as scene understanding, depth estimation, optical flow estimation, tracking, and visual grasping among others. Those new approaches have been slowly but surely closing the gap with traditional ones in terms of accuracy and efficiency, surpassing them in more and more cases and situations as those new techniques mature and better data is available. A key requirement for those new approaches to achieve outstanding accuracy while being able to generalize properly to unseen situations is a good dataset. Researchers need large-scale sets of images, which are representative enough for the problem at hand but at the same time include a considerable amount of variability, and complete ground truth for each one of them depending on the needs of the problem. Currently, there is a lack of such data since generating a dataset which satisfies those requirements is hard in practice due to the inherent difficulties of data collection. In this work, we focus on that challenge and aim to provide a unified benchmark for training, evaluating, and reproducing algorithms and methods on a wide spectrum of robotic vision problems.

The first matter that we must address is scale. The importance of large-scale data when working with data-hungry learning algorithms is critical in order to achieve proper results and generalization. This is due to the fact that collecting and labelling real-world data is a tedious and costly process. However, for a synthetic dataset to be useful, it must resemble reality as much as possible. In this regard, the use of photorealistic rendering solutions is a must. Other issues that we have to take into account are video-related intrinsics such as image resolution, frame rate, as well as the nature of each frame. Various works have highlighted the importance of high-resolution and high frame-rate data for various computer vision applications \cite{Handa2012}\cite{Held2016}. In that sense, our target is providing a high resolution dataset with fairly high frame rate (+60 FPS) and three data modalities for each frame: RGB-D, stereo, and 3D (colored point clouds) simulating the particularities of widespread sensors.

\begin{table*}[!t]
  \centering
  \caption{Summary of features of realistic/synthetic indoor datasets and environments.}
  \resizebox{\linewidth}{!}{
    \begin{tabular}{|r|c|c|c|c|c|c|c|c|c|c|c|c|c|}
      \hline
      & \multicolumn{2}{|c|}{\textbf{Scale}} & \multicolumn{2}{|c|}{\textbf{Photorealism}} & \multicolumn{2}{|c|}{\textbf{Video}} & \multicolumn{3}{|c|}{\textbf{Modalities}} & & & &\\
      \hline
      \textbf{Dataset} & \textbf{Frames} & \textbf{Layouts} & \textbf{Realism} & \textbf{Renderer} & \textbf{Seqs.} & \textbf{FPS} & \textbf{RGB} & \textbf{Depth} & \textbf{3D} & \textbf{Resolution} & \textbf{Interaction} & \textbf{Trajectories} & \textbf{Design}\\
      \hline
      \hline
      NYU-D V2 \cite{Silberman2012} & 1.5K & 464 & Real & - & - & - & $\bullet$ & $\bullet$ & - & $640\times480$ & - & - & Real\\
      Sun RGB-D \cite{Song2015} & 10K & N/A & Real & - & - & - & $\bullet$ & $\bullet$ & - & Mix & - & - & Real\\
      Stanford 2D-3D-S \cite{Armeni2017} & 70K & 270 & Real & - & - & - & $\bullet$ & $\bullet$ & $\bullet$ & $1080\times1080$ & - & - & Real\\
      Matterport 3D \cite{Chang2017} & 200K & 90 & Real & - & - & - & $\bullet$ & $\bullet$ & $\bullet$ & $1280\times1024$ & - & - & Real\\
      \hline
      \hline
      SunCG \cite{Song2016} & 130K & 45K & $\bullet\bullet$ & N/A & - & - & - & $\bullet$ & $\bullet$ & $640\times480$ & - & - & Manual\\ 
      PBR-Princeton \cite{Zhang2017} & 500K & 45K & $\bullet\bullet$ & Mitsuba & - & - & $\bullet$ & $\bullet$ & - & $640\times480$ & - & - & Manual\\
      SceneNet RGB-D \cite{McCormac2017} & 5M & 57 & $\bullet\bullet\bullet$ & NVIDIA Optix & 16895 & 1 & $\bullet$ & $\bullet$ & - & $320\times240$ & - & Synthetic & Random\\
      \textbf{Ours} & 8M & 16 & $\bullet\bullet\bullet\bullet\bullet$ & Unreal Engine & 512 & +60 & $\bullet$ & $\bullet$ & $\bullet$ & \textbf{$1920\times1080$} & \textbf{Hands} & \textbf{Realistic} & \textbf{Manual}\\      \hline
      \hline
      HoME \cite{Brodeur2017} & - & 45K & $\bullet\bullet$ & Panda3D & - & N/A & $\bullet$ & $\bullet$ & - & N/A & Physics & -  & Manual\\
      House3D \cite{Wu2018} & - & 45K & $\bullet\bullet$ & OpenGL & - & 600 & $\bullet$ & $\bullet$ & - & $120\times90$ & Physics & - & Manual\\
      AI2-THOR \cite{Kolve2017} & - & 120 & $\bullet\bullet\bullet\bullet$ & Unity & - & 13 & $\bullet$ & - & - & $300\times300$ & Actionable & - & Manual\\
      MINOS (Sun CG) \cite{Savva2017} & - & 45K & $\bullet\bullet$ & WebGL & - & 100 & $\bullet$ & $\bullet$ & - & N/A &  - & - & Manual\\
      MINOS (Matterport) \cite{Savva2017} & - & 90 & Real & - & - & 100 & $\bullet$ & $\bullet$ & - & N/A &  - & - & Real\\
    \hline
    \end{tabular}
  }
  \label{table:dataset_features}
\end{table*}

Although providing a large-scale and photorealistic dataset with decent ground truth is always highly useful for the community, there is a number of already existing works that do that but with certain shortcomings (low resolution, low frame rate, somewhat artificial scenes, or scarce modalities). Apart from iterating over those features to strengthen them, we wanted to include a set of novel ones which make our proposal truly stand out from the crowd: (1) hands, first person, and room points of view to make the dataset useful for various robotic applications, (2) realistic robot trajectories and head movements (if applicable) controlled by a human operator in \ac{VR}, and (3) visually realistic interactions with objects by using \ac{VR} controllers to move robotic hands with inverse kinematics. Furthermore, we release all the details and tools to allow researchers to generate custom data to fulfill their needs.

The rest of this paper is organized as follows. Firstly, Section \ref{sec:related_works} reviews the state of the art of each dataset-related features and ground truth information that we mentioned before. Next, Section \ref{sec:overview} provides an in-depth overview of the whole dataset and its generation. At last, Section \ref{sec:conclusion} draws conclusions about this work and proposes some future lines of research to improve it.

\section{RELATED WORKS}
\label{sec:related_works}

Synthetic image datasets have been used for a long time to benchmark vision algorithms \cite{Butler2012}. Recently, their importance has been highlighted for training and evaluating machine learning models for robotic vision problems \cite{Brodeur2017, Ros2016, Mahler2017dex}. Due to the increasing demand for annotated data, fostered by the rise of deep learning, real-world datasets are having a hard time to keep up since generating ground truth for them can be tedious and error-prone. Many indoor synthetic datasets have been proposed in the literature and some of them have been successfully used to train and validate deep architectures. In certain cases, it has even been proven that artificial data can be highly beneficial and increase the accuracy of state-of-the-art models on challenging real-world benchmarks and problems \cite{Shrivastava2017,Barbosa2018, Ros2016}. However, synthetic datasets have their own problems and existing ones can be improved in many ways. In this section, we review the most important aspects that make an indoor dataset suitable for training deep learning architectures: scale, photorealism, video, modalities, resolution, interactions, trajectories, and its design method. In addition, we also review the ground truth provided by them to determine their quality and which problems can be addressed by using that data. At last, we put our proposal in context by briefly enumerate its contributions and how our dataset improves and extends existing works.

\subsection{Features}

\begin{table*}[!t]
  \centering
  \caption{Overview of ground truth information provided by the reviewed datasets and environments.}
  \resizebox{\linewidth}{!}{
  \begin{tabular}{|r|c|c|c|c|c|c|c|c|c|c|}
    \hline
    \textbf{Dataset} & \textbf{2D BBox} & \textbf{2D Segm. Class} & \textbf{2D Segm. Inst.} & \textbf{3D BBox} & \textbf{3D Segm. Class} & \textbf{3D Segm. Inst.} & \textbf{3D Object Pose} & \textbf{Camera Pose} & \textbf{Hand Pose} & \textbf{Depth}\\
    \hline
    \hline
    NYU-D-V2 \cite{Silberman2012} & & $\bullet$ & $\bullet$ & & & & & $\sim$\footnotemark & & $\bullet$ \\
    Sun RGB-D \cite{Song2015} & & $\bullet$ & & $\bullet$ & & & & & & $\bullet$ \\
    Stanford 2D-3D-S \cite{Armeni2017} & & $\bullet$ & $\bullet$ & & $\bullet$ & $\bullet$ & & $\bullet$ & & $\bullet$\\
    Matterport 3D \cite{Chang2017} & & $\bullet$ & $\bullet$ & & $\bullet$ & $\bullet$ & & $\bullet$ & & $\bullet$\\
    \hline
    \hline
    Sun CG \cite{Song2016} & & & & & $\bullet$ & & & & & $\bullet$\\
    PBR-Princeton \cite{Zhang2017} & & $\bullet$ & $\bullet$ & & & & & & & $\bullet$\\
    SceneNet RGB-D \cite{McCormac2017} & & $\bullet$ & $\bullet$ & & & & $\bullet$ & $\bullet$ & & $\bullet$\\
    \textbf{Ours} & $\bullet$ & $\bullet$ & $\bullet$ & $\bullet$ & $\bullet$ & $\bullet$ & $\bullet$ & $\bullet$ (Multi) & $\bullet$ & $\bullet$ \\
    \hline
    \hline
    HoMe \cite{Brodeur2017} & & $\bullet$ & & & & & & & & $\bullet$ \\
    House3D \cite{Wu2018} & & $\bullet$ & & & & & & & & $\bullet$ \\
    AI2-THOR \cite{Kolve2017} & & & & & & & & & &\\ 
    MINOS (Sun CG) \cite{Savva2017} & $\bullet$ & $\bullet$ & $\bullet$ & & & & & & & $\bullet$ \\
    MINOS (Matterport) \cite{Savva2017} & $\bullet$ & $\bullet$ & $\bullet$ & & & & & & & $\bullet$ \\
    \hline
  \end{tabular}}
  \smallskip
  \label{table:dataset_ground_truth}
\end{table*}


The criteria that we used in this brief review of features (see Table \ref{table:dataset_features} for a summarized view of those features for the most popular and promising indoor datasets and environments that are already public) are the following ones:

\begin{itemize}
  \item \emph{Scale}: Data-driven algorithms such as deep learning approaches rely on massive amount of data to achieve unprecedented accuracy levels. Furthermore, massive amounts of information are not only needed to make those systems able to learn properly but also to give them the ability to generalize their knowledge to unseen situations. We measure scale according to the number of frames and possible layouts or room configurations (note that environments do not provide frames per-se so they are potentially infinite and that quantification makes no sense).
  \item \emph{Photorealism}: Recently, synthetic approaches have gained so much popularity since generating ground truth for them is an easy and automated task. Many successful stories have proven that artificial data can be used to train deep architectures for a wide variety of problems \cite{Ros2016,Lin2016,Mahendran2016,Jiang2017,Mueller2017,Zhang16}. Furthermore, some of them have highlighted the fact that training machine learning algorithms on virtual worlds even improves accuracy when they are applied to real-world scenarios \cite{Johnson-Roberson2016} \cite{Tobin2017}. In any case, the dataset will be more useful as it is closer to reality. We quantify realism on a scale of one to five according to the combination of texturing quality, rendering photorealism, object geometry, and layout coherence.
  \item \emph{Sequences}: Some problems can only be approached or at least they get much easier if video sequences are provided with a certain amount of \ac{FPS} is reached. For instance, object tracking mechanisms based on recurrent networks \cite{Held2016} or temporally coherent segmentation models \cite{Shelhamer2016} benefit from higher frame rates which provide smoother changes without huge differences between frames. In this regard, we report whether the dataset provides video data or not, the number of sequences and the average framerate; for the environments the framerate indicates how many actions/renderings can be performed per frame.
  \item \emph{Modalities}: There is certain importance in providing as many data modalities as possible. On the one hand, some problems can only be addressed if data from a particular nature is available, e.g., depth information is needed to make a system learn to estimate depth in a supervised manner. On the other hand, even if a problem can be solved using a concrete data modality, having more sources of information available might foster the development of alternative ways of fusing that extra data to boost performance. For all the reviewed datasets, we report the kind of data modalities that they provide.
  \item \emph{Resolution}: Vision systems usually benefit from larger image sizes since higher resolution means better image quality and thus more robust features to be learned. One notorious success case of high-resolution imaging in deep learning is the case of Baidu Vision's system \cite{Wu2015} which introduced a novel multi-scale and high-resolution model to achieve the best score on the well-known ImageNet challenge \cite{Russakovsky2015} by the time they published the work. However, this is not always true and for some applications it is important to find balance in the tradeoff between accuracy and performance when processing large images. We indicate the image resolution for each dataset and environment.
  \item \emph{Interaction}: Despite the importance of hand pose estimation and object interaction in many applications, e.g., grasping and dexterous manipulation, this aspect is often neglected. The main reason about this scarcity is the difficulty of generating annotations for the hand joints and moving objects. In some cases, interaction is reduced to simple actions with binary states. We report the kind of interaction that is performed on each dataset (physics simulations for collisions, actionable items, or full object interaction with robotic hands).
  \item \emph{Trajectories}: For sequences, the way trajectories are generated plays an important role in the quality or applicability of the dataset. In order to make the dataset distribution as close as possible to the application scenario one, the camera trajectory should be similar too. Real-world datasets usually leverage handheld or head-mounted devices to generate human-like trajectories and, in the case of a robot, capture devices are usually mounted in the same place where they will work when deployed. Synthetic datasets must devise strategies to place and orient cameras in a coherent manner. For each dataset that provides video sequences, we report the way those trajectories are generated.
  \item \emph{Design}: The design of scene layouts is another factor that must be taken into account to make the dataset as similar as possible to the real-world. This means that scenes must be coherent, i.e., objects and lights must be positioned according to actual room layouts. Generating coherent rooms with plausible configurations synthetically to achieve large-scale data is a hard task and only \emph{SceneNet}\cite{Handa2015} and \emph{SceneNet RGB-D}\cite{McCormac2017} approached the room design problem algorithmically; their results are plausible but oftentimes not really representative of what a real-world room would look like due to artificial object positioning and orientations. For this feature, we report whether the design is real, manual or algorithmic/synthetic.
\end{itemize}

\subsection{Ground Truth}

Although all the aforementioned features are of utmost importance for a dataset, ground truth is the cornerstone that will dictate the usefulness of the data. It determines the problems that can be solved by using the available data.  Table \ref{table:dataset_ground_truth} shows the ground truth information provided by each one of the reviewed datasets including ours, which completes and offers more annotations than the state of the art.

\footnotetext{It provides Roll, Yaw, Pitch and Tilt angle of the device from an accelerometer.}

\subsection{Our Proposal in Context}

After analyzing the strong points and weaknesses of the most popular indoor datasets, we aimed to combine the strengths of all of them while addressing their weaknesses and introducing new features. The major contributions of our novel dataset with regard to the current state of the art are:

\begin{itemize}
	\item \textbf{Large-scale and high level of photorealism.}
  \item \textbf{High frame-rate and resolution sequences.}
  \item \textbf{Multiple data modalities (RGB-D/3D/Stereo).}
	\item \textbf{Realistic robot trajectories with multiple \acsp{PoV}.}
	\item \textbf{Robot interaction within the scene.}
  \item \textbf{Ground truth for many vision problems.}
  \item \textbf{Open-source pipeline and tools\footnote{\url{https://github.com/3dperceptionlab/therobotrix}}.}
\end{itemize}

\section{OVERVIEW}
\label{sec:overview}

After reviewing the existing datasets and stating the contributions of our proposal, we will provide a detailed description of its main features and how did we achieve them: hyper-photorealism by combining a powerful rendering engine with extremely detailed scenes and realistic robot movements and trajectories. We will also provide an overview of the data collection pipeline: the online sequence recording procedure and the offline data and ground truth generation process. Furthermore, we will list the contents of the dataset and the tools that we provide as open-source software for the robotics research community.

\subsection{Photorealistic Rendering}

The rendering engine we chose to generate photorealistic RGB images is \acf{UE4}\footnote{\url{http://www.unrealengine.com}}. The reasons for this choice are the following ones: (1) it is arguably one of the best game engines able to produce extremely realistic renderings, (2) beyond gaming, it has become widely adopted by VR developers and architectural visualization experts so a whole lot of tools, examples, documentation, and assets are available; (3) due to its impact, many hardware solutions offer plugins for \ac{UE4} that make them work out-of-the-box; and (4) Epic Games provides the full C++ source code and updates to it so the full suite can be easily used and modified.

\begin{figure}[!hbt]
  \centering
  \includegraphics[width=0.9\linewidth]{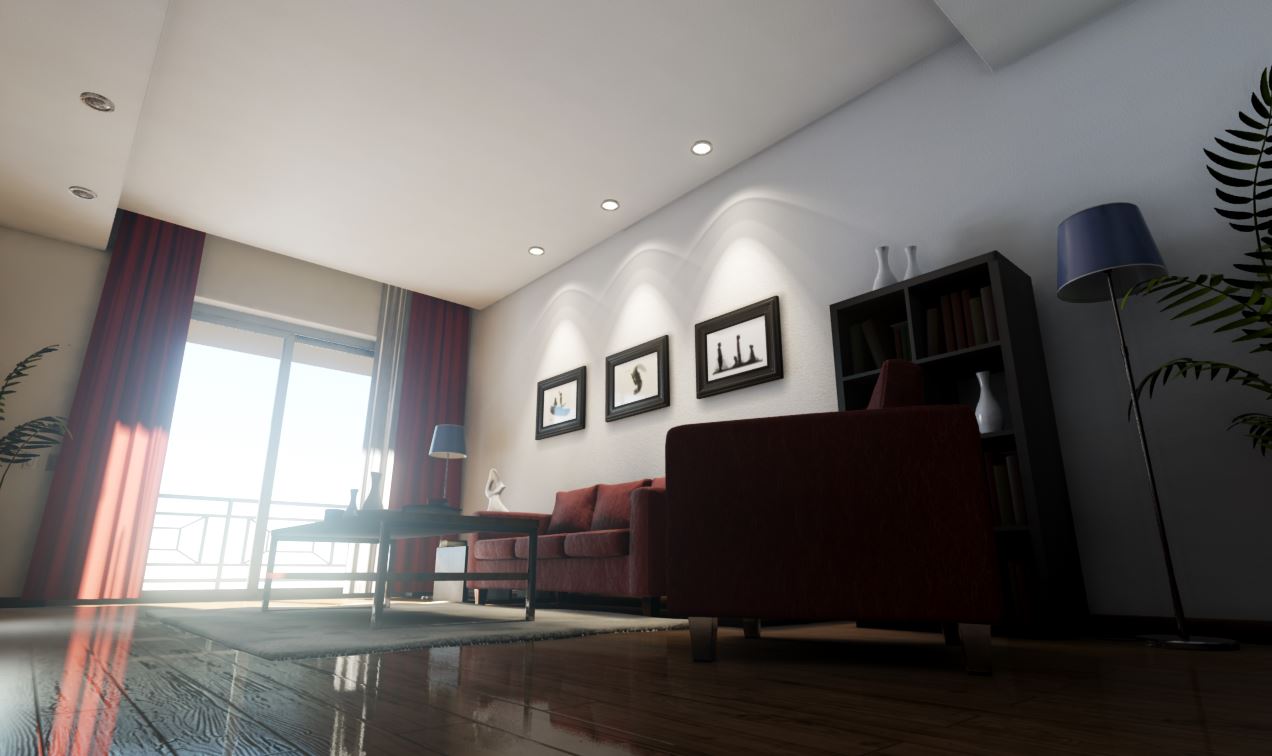}
  \caption{Snapshot of the daylight room setup for the \emph{Realistic Rendering} released by Epic Games to show off the rendering capabilities of \ac{UE4}.}
  \label{fig:realistic_rendering}
\end{figure}

Arguably, the most attractive feature of \ac{UE4} that made us take that decision is its capability to render photorealistic scenes like the one shown in Figure \ref{fig:realistic_rendering} in real time. Some \ac{UE4} features that enable this realism are: physically-based materials, pre-calculated bounce light via Lightmass, stationary lights using IES profiles, post-processing, and reflections.

It is also important to remark that we do have strict real-time constraints for rendering during sequence recording since we need to immerse a human agent in virtual reality to record the sequences. \ac{UE4} is engineered for virtual reality with a specific rendering solution for it named \emph{Forward Renderer}. That renderer is able to generate images that meet our quality standards at 90 \ac{FPS} thanks to high-quality lighting features, \ac{MSAA}, and instanced stereo rendering.

\subsection{Scenes}

\begin{figure}[!th]
  \centering
  \includegraphics[width=\linewidth]{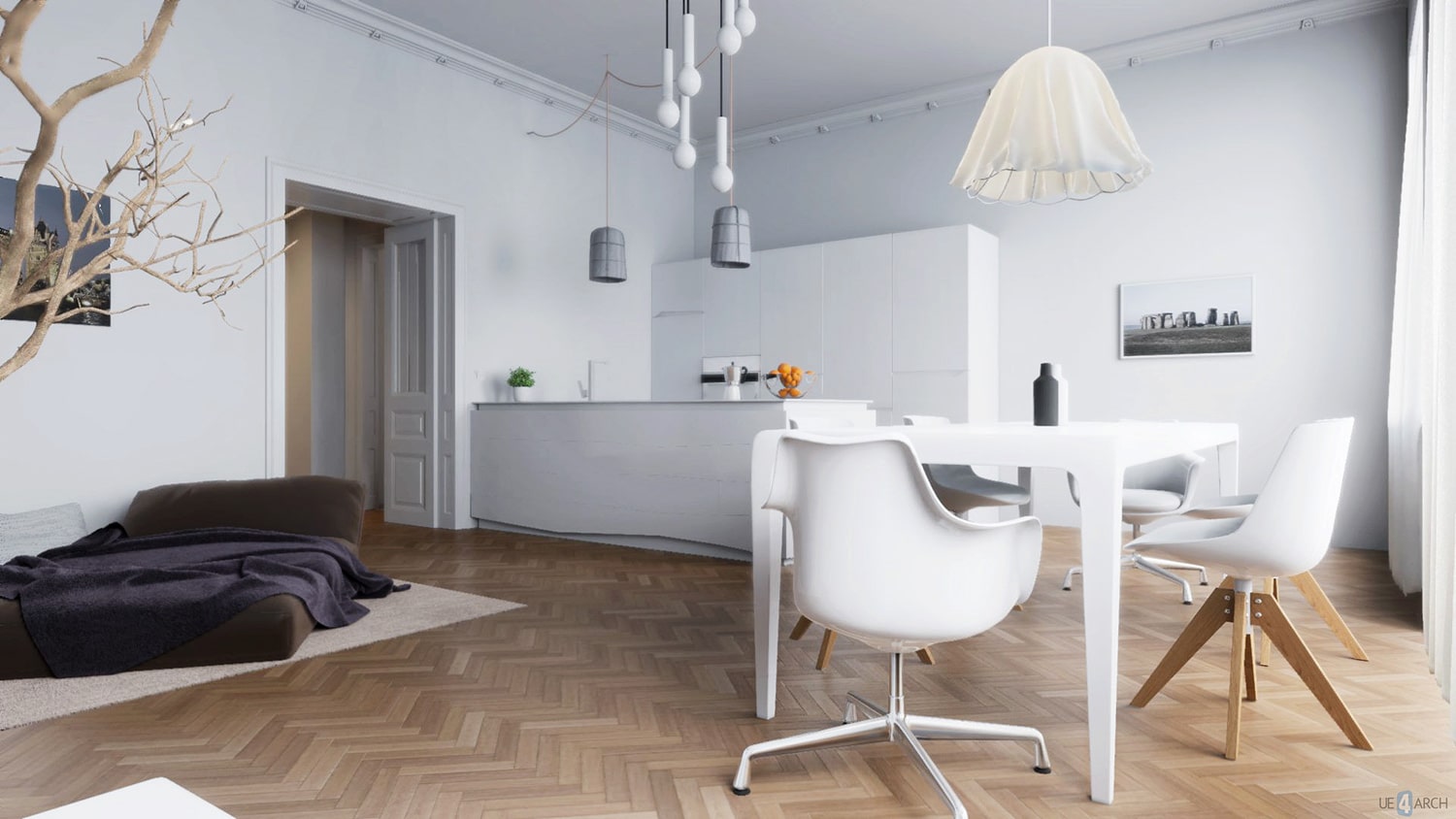}
  \caption{Viennese Apartment archviz project snapshot by UE4Arch.}
  \label{fig:ue4arch}
\end{figure}

To complement the rendering engine and achieve the desired level of photorealism we needed coherent indoor scenes with extreme attention to detail. UE4Arch \footnote{\url{https://ue4arch.com/}} is a company devoted to creating hyper-realistic and real-time architecture visualizations with \ac{UE4}. We take advantage of various house projects and assets created by that company to populate our dataset with rich environments and layouts. Figure \ref{fig:ue4arch} shows a sample project from UE4Arch.

\subsection{Robot Integration}

Seamlessly integrating robots in our scenes and making them controllable in \ac{VR} by a human agent to record sequences requires three issues to be solved: (1) gaze and head movement with first person \ac{PoV}, (2) inverse kinematics to be able to move them with motion controllers and reach for objects, and (3) locomotion to displace the robot within the scene.

\begin{figure}[!htb]
  \centering
  \includegraphics[width=0.49\linewidth]{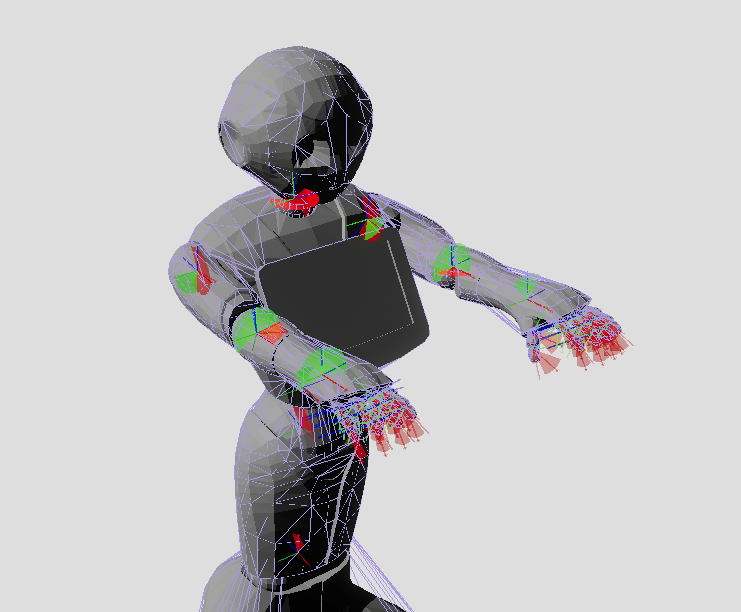}
  \includegraphics[width=0.49\linewidth]{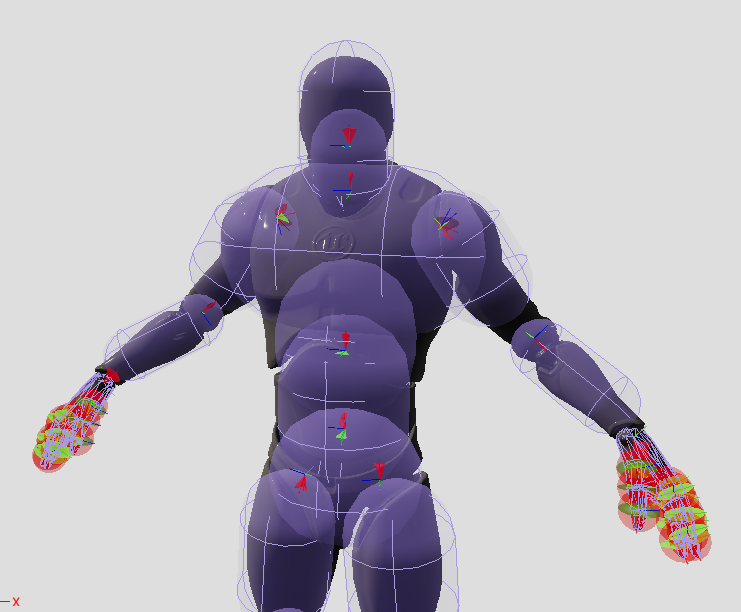}
  \caption{Pepper and Mannequin integrated with colliders and constraints.}
  \label{fig:robot_integration}
\end{figure}

The first issue is solved by using the Oculus Rift headset to control the robot's head movement and render its first person \ac{PoV}. Inverse kinematics for the virtual robot are manually implemented with \ac{FABRIK}, a built-in inverse kinematics solver in \ac{UE4} that works on a chain of bones of arbitrary length. Locomotion is handled by thumbsticks on the Oculus Touch  motion controllers. By doing this we are able to integrate any robot in FBX format such as Unreal's mannequin or the well-known Pepper by Aldebaran (see Figure \ref{fig:robot_integration}).

\clearpage

\subsection{Object Interaction}

One key aspect of the dataset is the simulation of realistic interactions with objects. On the one hand, we need to simulate basic physics to move, push, pull, or lift objects in the scene. On the other hand, for small props we need to provide a way to grasp them if they fit in the robot's hand for a more complex interaction.

\begin{figure}[!htb]
  \centering
  \includegraphics[width=0.9\linewidth]{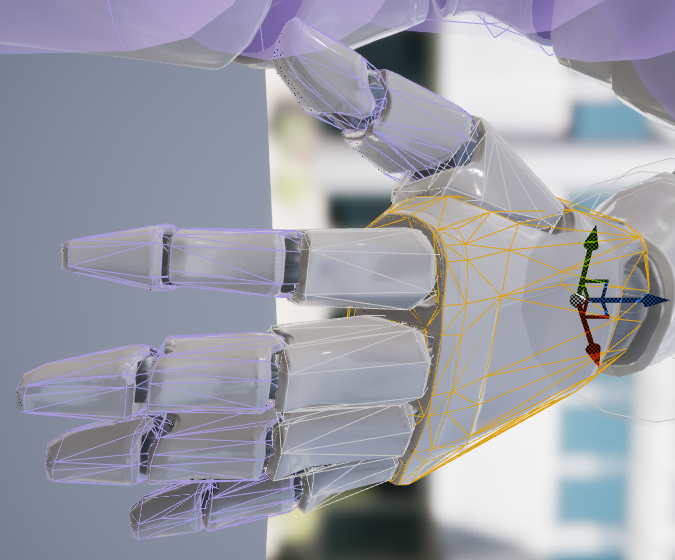}\\
  \includegraphics[width=0.3\linewidth]{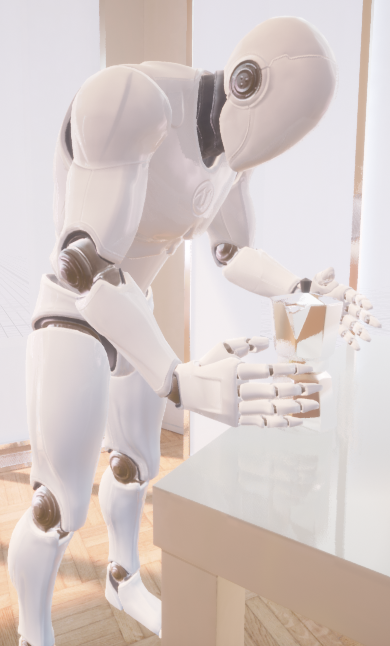}
  \includegraphics[width=0.3\linewidth]{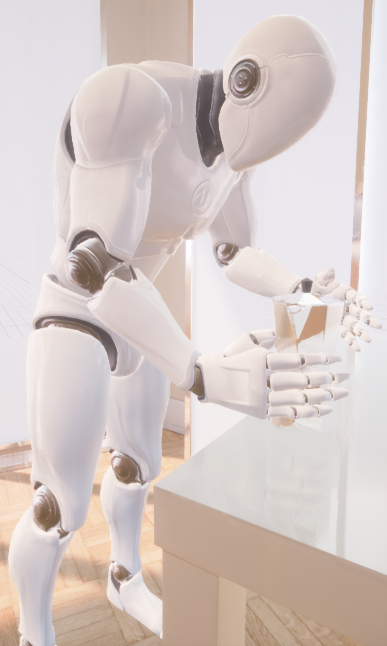}
  \includegraphics[width=0.3\linewidth]{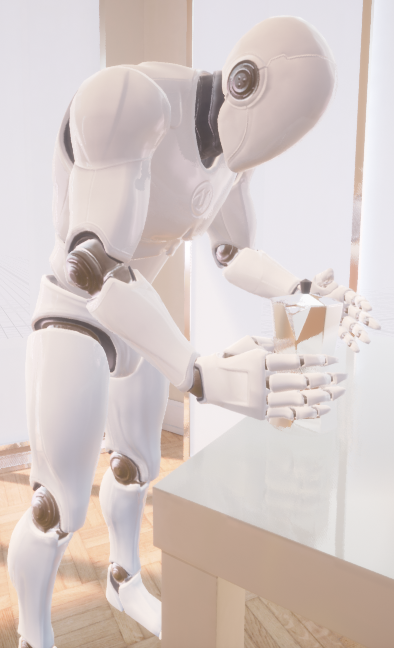}
  \caption{Single convex hull colliders for \ac{UE4}'s mannequin robotic hand and a example of object interaction and grasping.}
  \label{fig:object_interaction}
\end{figure}

To solve this issue we leverage \ac{UE4}'s built-in physics engine. Complex surfaces such as the robot's hands or its body and certain objects are modeled with single convex hulls as collider primitives. Simpler geometries are just approximated with less sophisticated primitives such as spheres or boxes. Those objects which are susceptible of being manipulated are properly weighted to resemble their real-world physical behavior. Furthermore, we implemented a simple yet visually appealing grasping approach to ease object interaction and make it look as realistic as possible: firstly, each hand is animated to have a pose blending from open to closed (this blending or interpolation is controlled by the analog triggers from each Oculus Touch); second, for each object that is small enough to be grabbed, we check which hand bones collide with it; if the five fingers and the palm collide with the object, we attach that object to the hand and stop simulating its physics so that it is stable in the hand; once those collisions are no longer happening, the object is detached and its physics are enabled again. Figure \ref{fig:object_interaction} shows the hand colliders and examples of interaction and grasping.

\subsection{Sequence Recording and Data Collection}

To record and generate all the data for this dataset, we made extensive use of a tool that was specifically built for this dataset: UnrealROX \cite{Martinez-Gonzalez2018}, a virtual reality environment for generating synthetic data for various robotic vision tasks. In such environment, a human operator can be embodied, in virtual reality, as a robot agent inside a scene to freely navigate and interact with objects as if it was a real-world robot. Our environment is built on top of \ac{UE4} to take advantage of its advanced \ac{VR}, rendering, and physics capabilities. That system provides the following features: (1) a visually plausible grasping system for robot manipulation which is modular enough to be applied to various finger configurations, (2) routines for controlling robotic hands and bodies with commercial \ac{VR} setups such as Oculus  Rift and HTC Vive Pro, (3) a sequence recorder component to store all the information about the scene, robot, and cameras while the human operator is embodied as a robot, (4) a sequence playback component to reproduce the previously recorded sequence offline to generate raw data such as RGB, depth, normals,  or  instance  segmentation  images,  (5)  a  multi-camera  component  to  ease  the  camera  placement  process and  enable  the  user  to  attach  them  to  specific  robot  joints and  configure  their  parameters  (resolution,  noise  model, field of view), and (6) open-source code, assets, and tutorials for all those components and other subsystems that tie them together.

In order to generate the dataset, we first collect data in an online and interactive manner by immersing human agents in the virtual indoor environment so that they can freely move, look, and interact with the scene (respecting robot constraints). In this stage, we use UnrealROX with \ac{UE4} to render our scene into an Oculus/HTC Vive Pro \ac{VR} headset worn by a person equipped with motion controllers for hand movement. During this phase, we gather all the information that we would need to replay the whole sequence offline to collect data and generate annotations without lagging the rendering process.

This data collection process is performed by an actor in \ac{UE4} which \emph{ticks} on every rendered frame and asynchronously dumps to a text file the $SE(3)$ pose (location and rotation) of every object and camera in the scene. It also dumps the full pose for each bone of the robot. This text file just contains a timestamp for each frame and the aforementioned raw information to have a minimal impact on performance. In fact, this process allows us to render at 80+ \ac{FPS} thanks to asynchronous and threaded writes to files. After the whole sequence is recorded, the text file is converted to JSON for better interpretability. Figure \ref{fig:sequence_recording} shows a diagram of this sequence recording procedure.

\begin{figure}[!hbt]
  \centering
  \includegraphics[width=\linewidth]{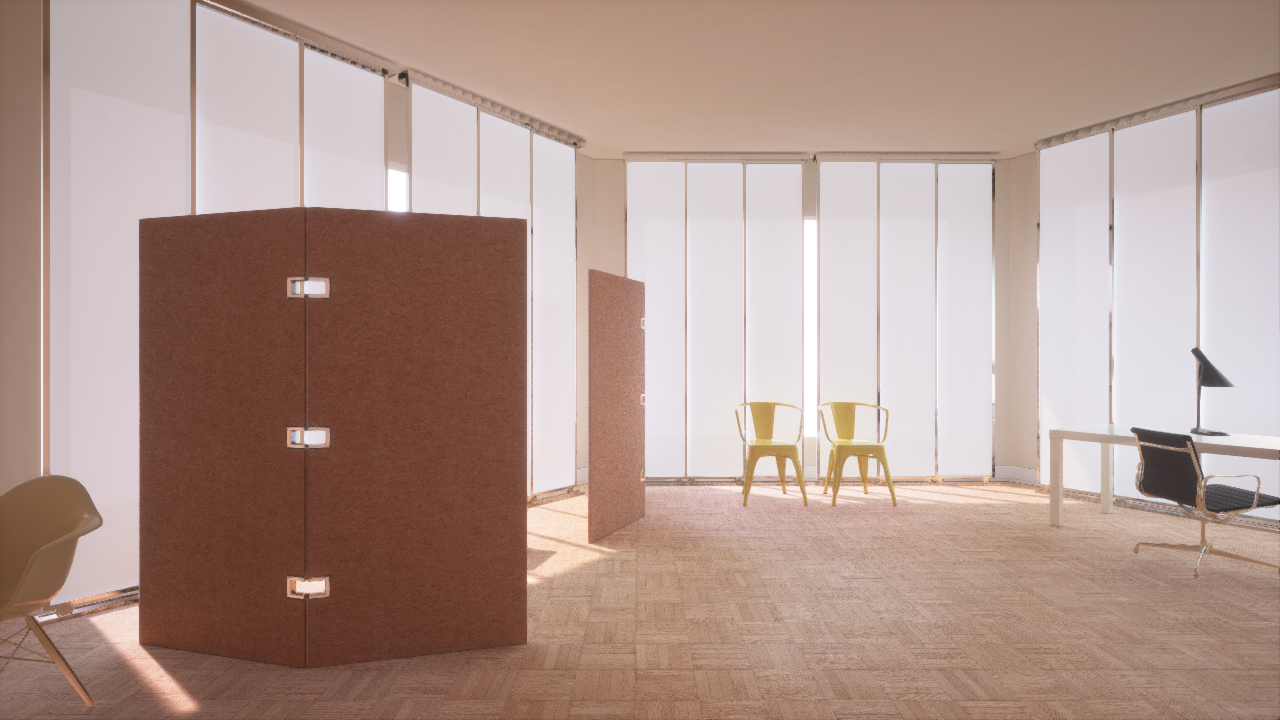}\\
  \smallskip
  \includegraphics[width=0.4935\linewidth]{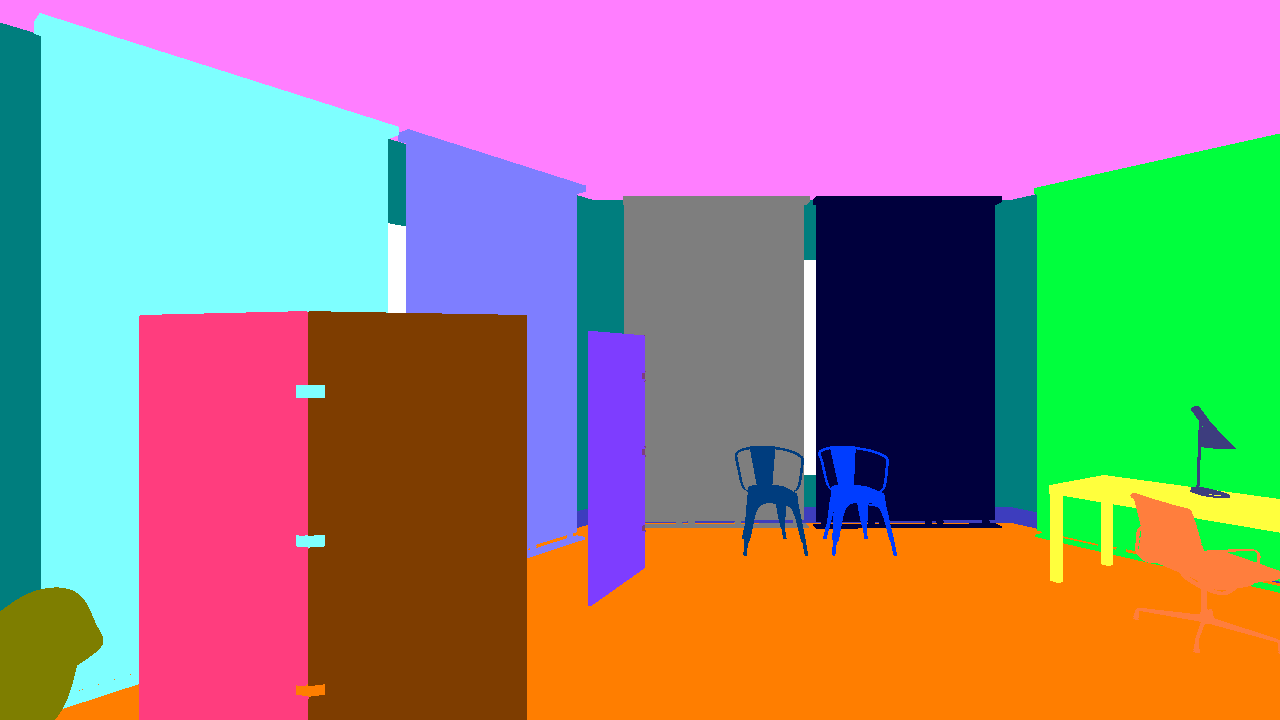}
  \includegraphics[width=0.4935\linewidth]{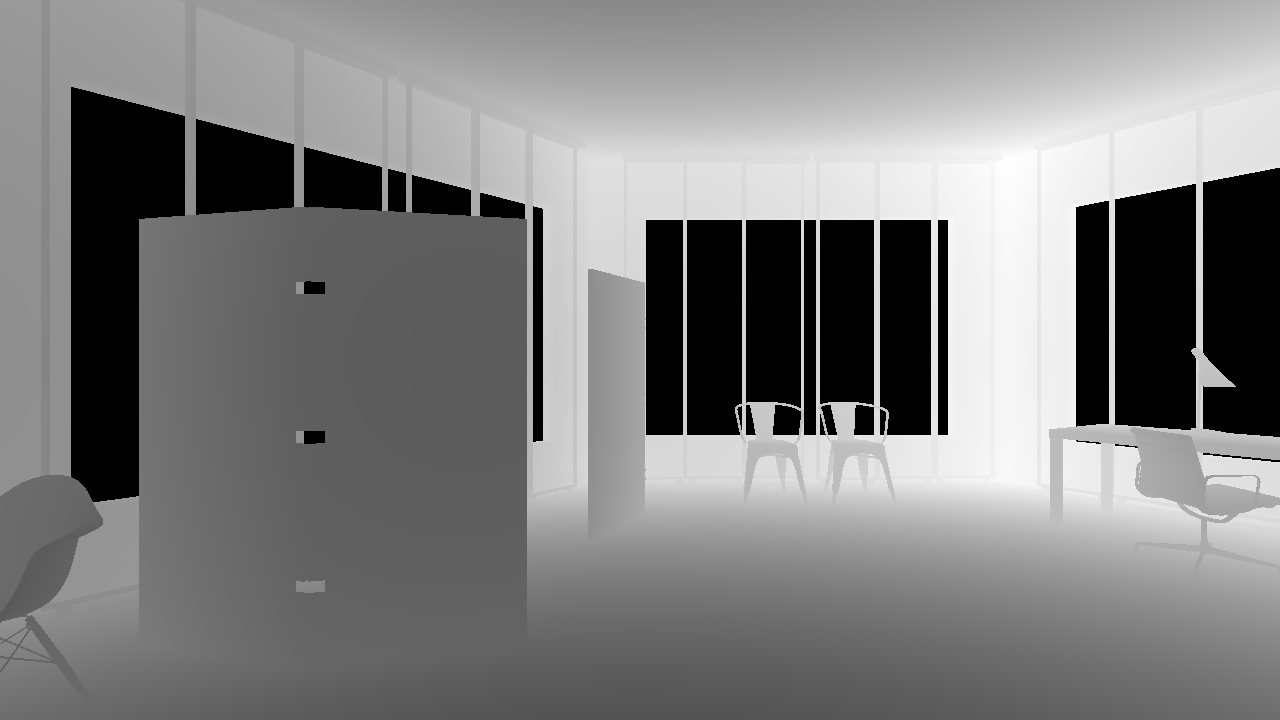}
  \caption{RGB, instance segmentation masks, and depth map generated with UnrealROX for a synthetic scene rendered in \acf{UE4}.}
  \label{fig:unrealcv_overview}
\end{figure}

After that we use the playback component of UnrealROX to reproduce sequences frame by frame by setting the corresponding poses for the robot and all objects and cameras. Once the frame is exactly as it was when we recorded the sequence, we just generate all the needed data offline by requesting the appropriate information such as RGB images, depth maps, or segmentation masks through the interface. Figure \ref{fig:unrealcv_overview} shows some examples of raw data generated with UnrealROX.

\subsection{Ground Truth Generation}

\begin{figure}[!htb]
  \centering
  \includegraphics[width=0.325\linewidth]{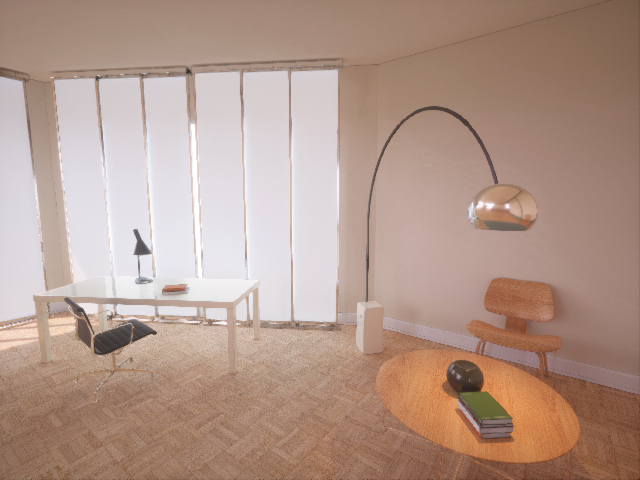}
  \includegraphics[width=0.325\linewidth]{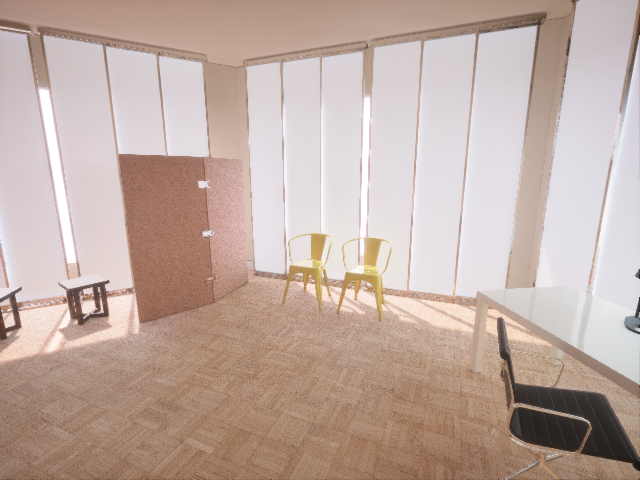}
  \includegraphics[width=0.325\linewidth]{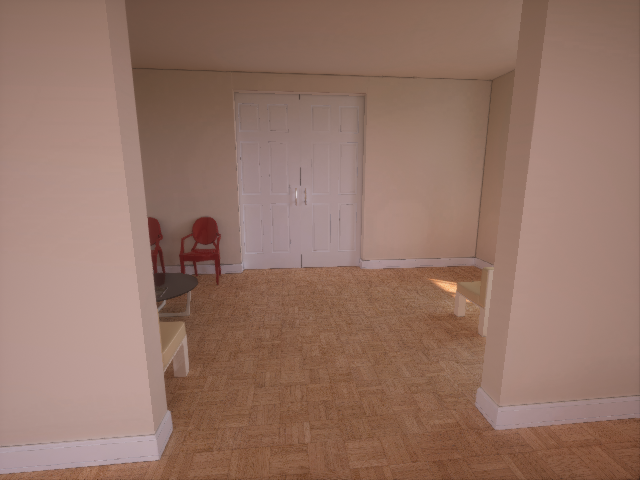}\\
  \smallskip
  \includegraphics[width=0.325\linewidth]{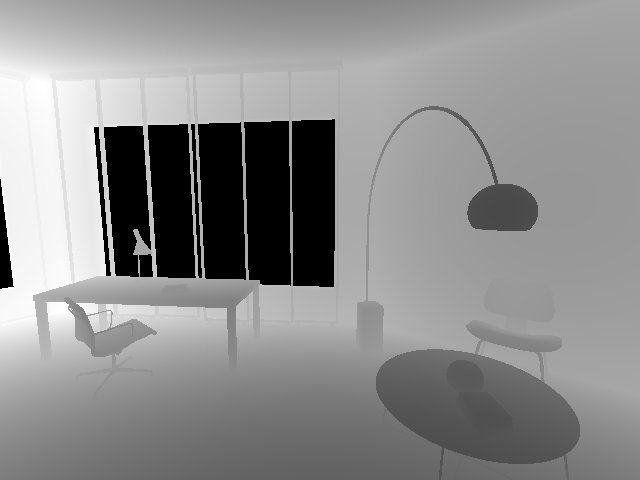}
  \includegraphics[width=0.325\linewidth]{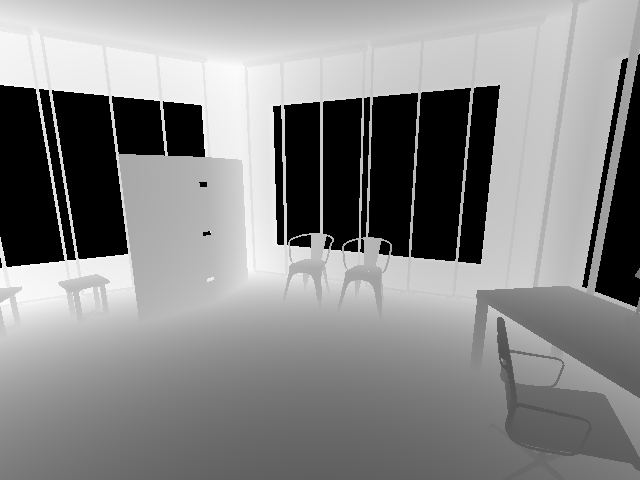}
  \includegraphics[width=0.325\linewidth]{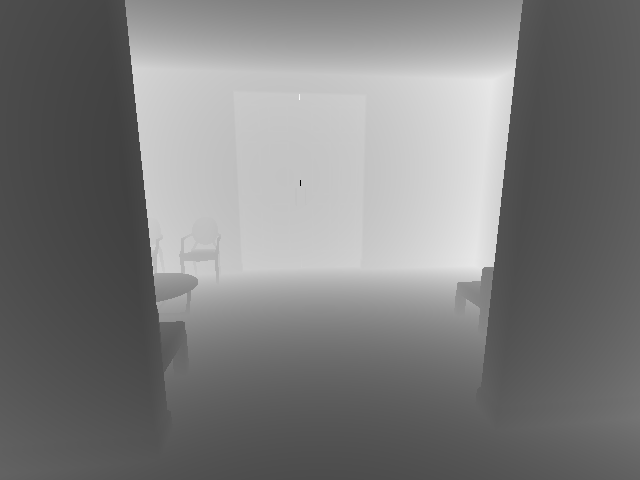}\\
  \smallskip
  \includegraphics[width=0.325\linewidth]{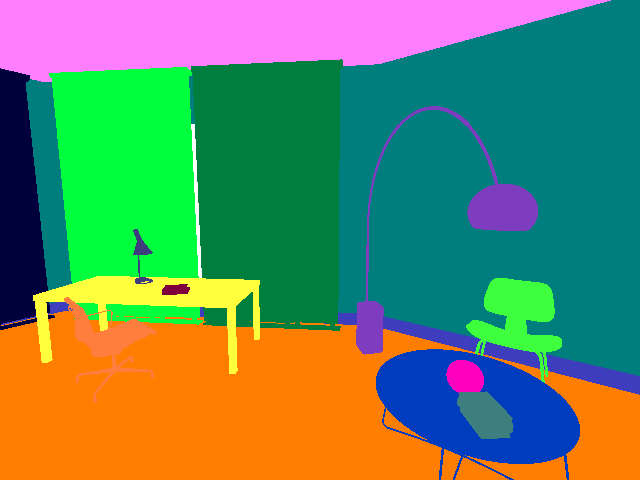}
  \includegraphics[width=0.325\linewidth]{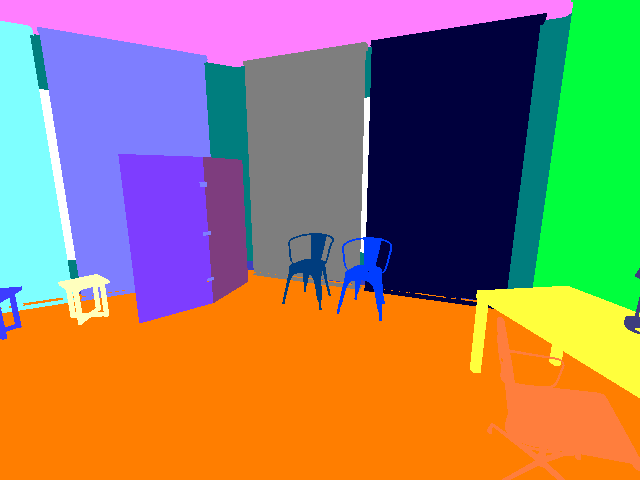}
  \includegraphics[width=0.325\linewidth]{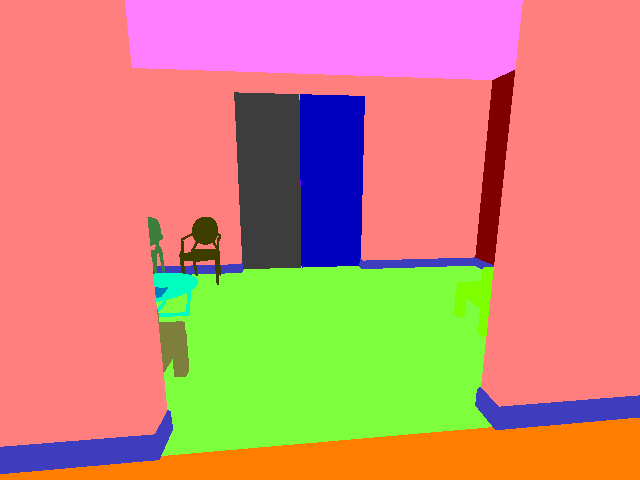}\\
  \smallskip
  \includegraphics[width=0.325\linewidth]{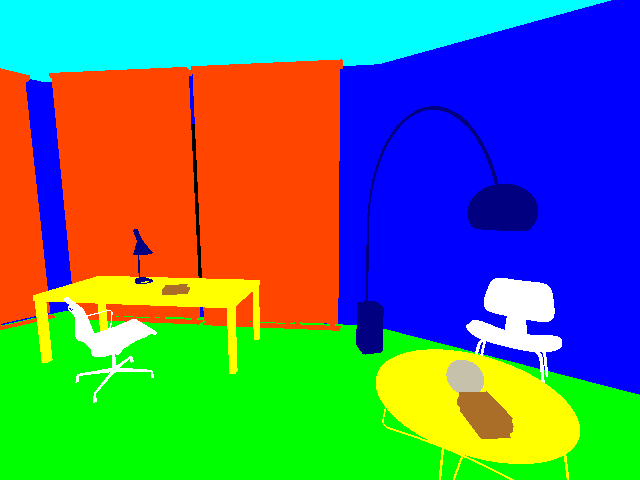}
  \includegraphics[width=0.325\linewidth]{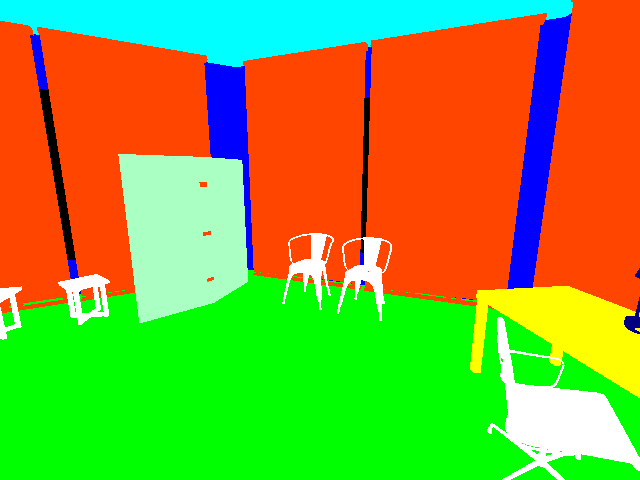}
  \includegraphics[width=0.325\linewidth]{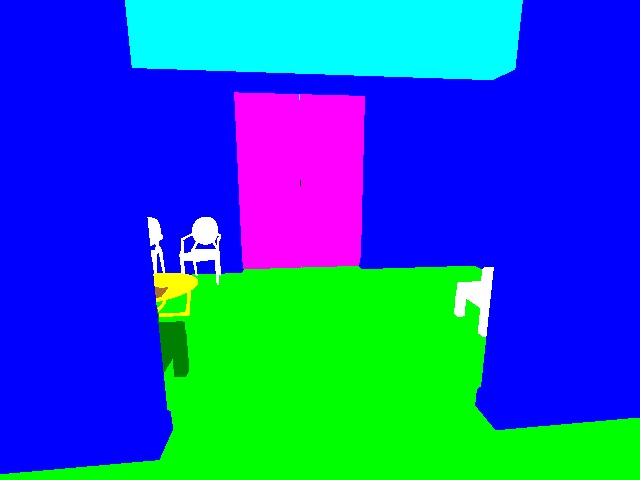}\\
  \smallskip
  \includegraphics[width=0.325\linewidth]{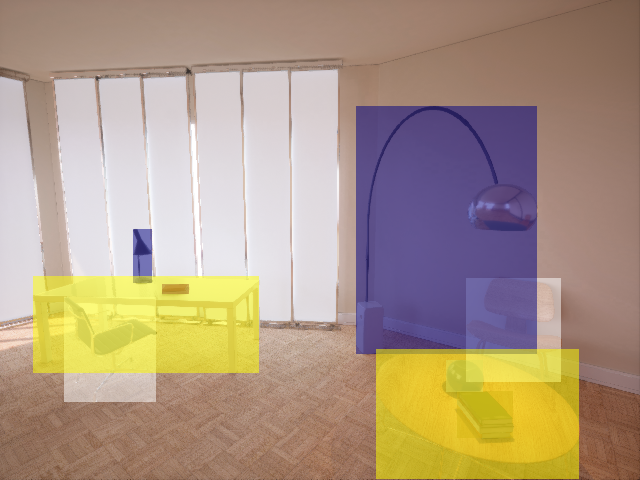}
  \includegraphics[width=0.325\linewidth]{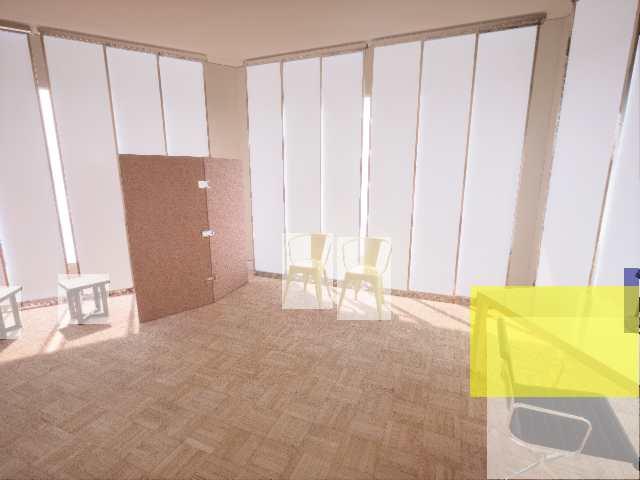}
  \includegraphics[width=0.325\linewidth]{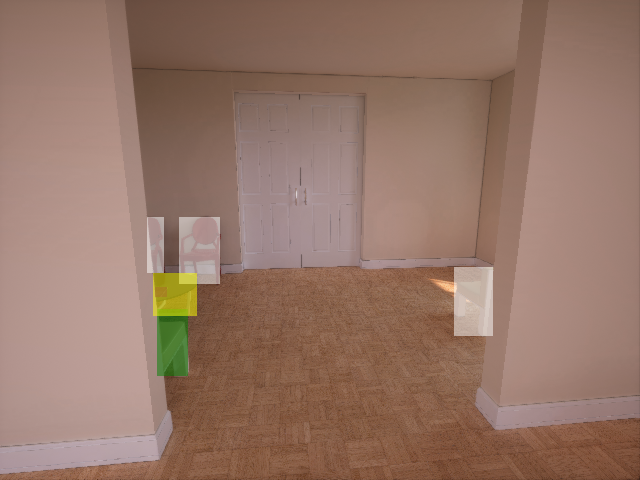}\\
  \caption{Ground truth generation examples (from top to bottom: RGB, depth, instance masks, class masks, and bounding boxes).}
  \label{fig:ground_truth}
\end{figure}

After collecting RGB, depth, and instance segmentation masks for each frame of a sequence, we can use that data to generate annotations. In our dataset release, we only include that raw data from the sequence and certain configuration files generated by the client in order to be able to produce ground truth annotations offline and on demand. We decoupled the data collection and the ground truth generation processes for a simple reason: practicality. In this way, to use the dataset researchers only need to download the RGB, depth, instance segmentation masks, and the generator code to locally generate whichever annotation their problems require in the appropriate format instead of fetching the full bundle in a predefined one. The generator takes that raw data and additional information generated by the client (camera configuration, object classes, colors, and instance mapping) and outputs 2D/3D bounding boxes in VOC format, point clouds, 2D/3D class segmentation masks, and 3D instance segmentation masks (hand pose and camera pose information is embedded in the sequence recording). Figure \ref{fig:ground_truth} shows some examples of ground truth generation.

\subsection{Content}

\begin{table*}[!tb]
  \centering
  \caption{Classes for semantic segmentation and object detection.}
  \resizebox{\linewidth}{!}{
  \begin{tabular}{|c|c|c|c|c|c|c|c|c|c|c|c|c|c|c|}
    \hline
    & \textbf{0}  & \textbf{1} & \textbf{2} & \textbf{3} & \textbf{4} & \textbf{5} & \textbf{6} & \textbf{7} & \textbf{8} & \textbf{9} & \textbf{10} & \textbf{11} & \textbf{12}\\
    \textbf{Type} & \textcolor{rgb0}{\rule{0.75cm}{0.25cm}}& \textcolor{rgb1}{\rule{0.75cm}{0.25cm}}& \textcolor{rgb2}{\rule{0.75cm}{0.25cm}}& \textcolor{rgb3}{\rule{0.75cm}{0.25cm}}& \textcolor{rgb4}{\rule{0.75cm}{0.25cm}}& \textcolor{rgb5}{\rule{0.75cm}{0.25cm}}& \textcolor{rgb6}{\rule{0.75cm}{0.25cm}}& \textcolor{rgb7}{\rule{0.75cm}{0.25cm}}& \textcolor{rgb8}{\rule{0.75cm}{0.25cm}}& \textcolor{rgb9}{\rule{0.75cm}{0.25cm}}& \textcolor{rgb10}{\rule{0.75cm}{0.25cm}}& \textcolor{rgb11}{\rule{0.75cm}{0.25cm}}& \textcolor{rgb12}{\rule{0.75cm}{0.25cm}}\\
    \hline
    Semantic & void & wall & floor & ceiling & window & door & table & chair & lamp & sofa & cupboard & screen & robot\\
    Detection & - & - & - & - & - & - & table & chair & lamp & sofa & cupboard & screen & robot\\
    \hline
    & \textbf{13}  & \textbf{14} & \textbf{15} & \textbf{16} & \textbf{17} & \textbf{18} & \textbf{19} & \textbf{20} & \textbf{21} & \textbf{22} & \textbf{23} & \textbf{24} & \textbf{25}\\
    \textbf{Type} & \textcolor{rgb13}{\rule{0.75cm}{0.25cm}}& \textcolor{rgb14}{\rule{0.75cm}{0.25cm}}& \textcolor{rgb15}{\rule{0.75cm}{0.25cm}}& \textcolor{rgb16}{\rule{0.75cm}{0.25cm}}& \textcolor{rgb17}{\rule{0.75cm}{0.25cm}}& \textcolor{rgb18}{\rule{0.75cm}{0.25cm}}& \textcolor{rgb19}{\rule{0.75cm}{0.25cm}}& \textcolor{rgb20}{\rule{0.75cm}{0.25cm}}& \textcolor{rgb21}{\rule{0.75cm}{0.25cm}}& \textcolor{rgb22}{\rule{0.75cm}{0.25cm}}& \textcolor{rgb23}{\rule{0.75cm}{0.25cm}}& \textcolor{rgb24}{\rule{0.75cm}{0.25cm}}& \textcolor{rgb25}{\rule{0.75cm}{0.25cm}}\\
    \hline
    Semantic & frame & bed & fridge & whiteboard & book & bottle & plant & furniture & toilet & phone & bathtub & cup & mat\\
    Detection & frame & bed & fridge & whiteboard & book & bottle & plant & - & toilet & phone & bathtub & cup & mat\\
    \hline
    & \textbf{26} & \textbf{27} & \textbf{28} & \textbf{29} & \textbf{30} & \textbf{31} & \textbf{32} & \textbf{33} & \textbf{34} & \textbf{35} & \textbf{36} & \textbf{37} & \textbf{38}\\
    \textbf{Type} & \textcolor{rgb26}{\rule{0.75cm}{0.25cm}}& \textcolor{rgb27}{\rule{0.75cm}{0.25cm}}& \textcolor{rgb28}{\rule{0.75cm}{0.25cm}}& \textcolor{rgb29}{\rule{0.75cm}{0.25cm}}& \textcolor{rgb30}{\rule{0.75cm}{0.25cm}}& \textcolor{rgb31}{\rule{0.75cm}{0.25cm}}& \textcolor{rgb32}{\rule{0.75cm}{0.25cm}}& \textcolor{rgb33}{\rule{0.75cm}{0.25cm}}& \textcolor{rgb34}{\rule{0.75cm}{0.25cm}}& \textcolor{rgb35}{\rule{0.75cm}{0.25cm}}& \textcolor{rgb36}{\rule{0.75cm}{0.25cm}}& \textcolor{rgb37}{\rule{0.75cm}{0.25cm}}& \textcolor{rgb38}{\rule{0.75cm}{0.25cm}}\\
    \hline
    Semantic & mirror & sink & box & mouse & keyboard & bin & cushion & shelf & bag & curtain & kitchen\_stuff & bath\_stuff & prop\\
    Detection & mirror & sink & box & mouse & keyboard & bin & cushion & shelf & bag & - & kitchen\_stuff & bath\_stuff & prop\\
    \hline
  \end{tabular}}
  \label{table:classes}
\end{table*}

Using this pipeline, we generated a dataset of $512$ sequences recorded on $16$ room layouts (some samples are shown in Figure \ref{fig:scenes}) at $+60$ \ac{FPS} with a duration that spans between one and five minutes. That means a total of approximately $8$ million individual frames. For each one of those frames we provide the following data:

\begin{itemize}
  \item 3D poses for the cameras, objects, and joints.
  \item RGB image @ $1920\times1080$ in JPEG.
  \item Depth map @ $1920\times1080$ in 16-bit PNG.
  \item 2D instance mask @ $1920\times1080$ in 24-bit PNG.
\end{itemize}

And also annotations:

\begin{figure}[!tb]
  \centering
  \includegraphics[width=0.325\linewidth]{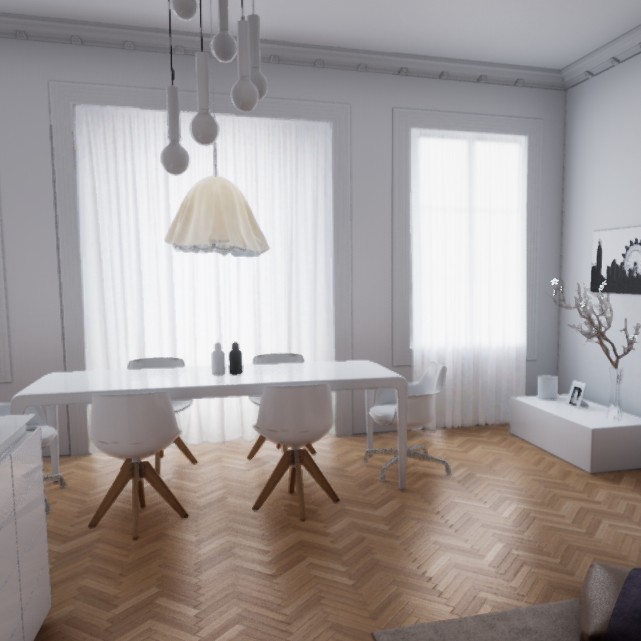}
  \includegraphics[width=0.325\linewidth]{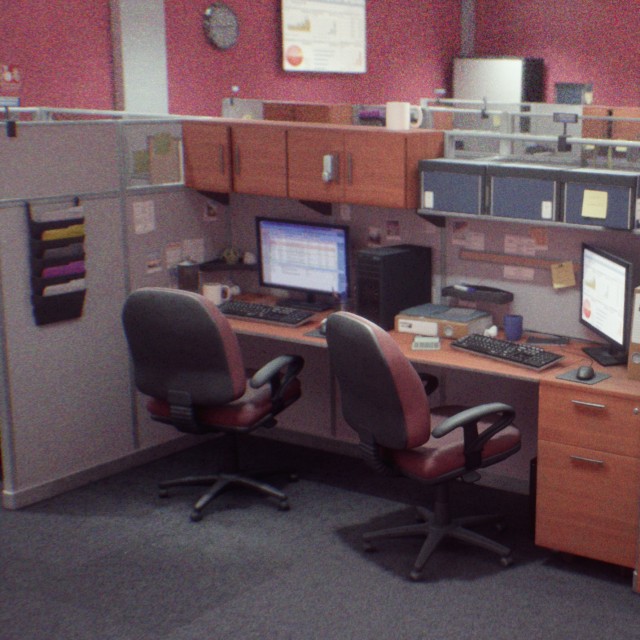}
  \includegraphics[width=0.325\linewidth]{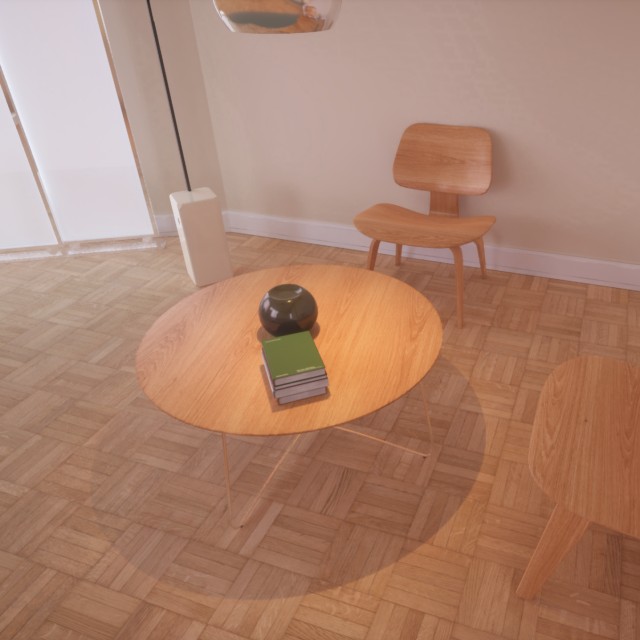}\\
  \smallskip
  \includegraphics[width=0.325\linewidth]{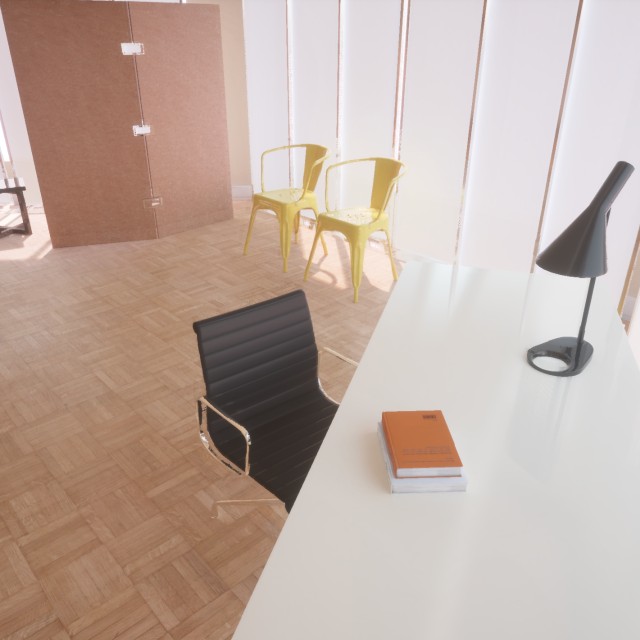}
  \includegraphics[width=0.325\linewidth]{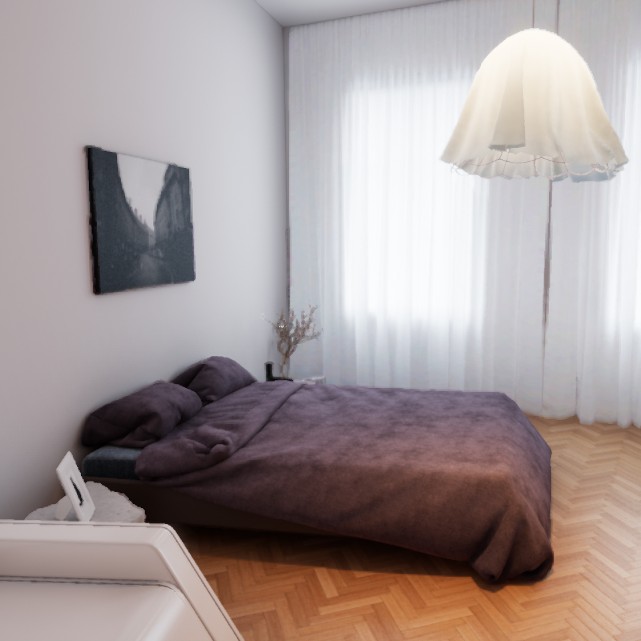}
  \includegraphics[width=0.325\linewidth]{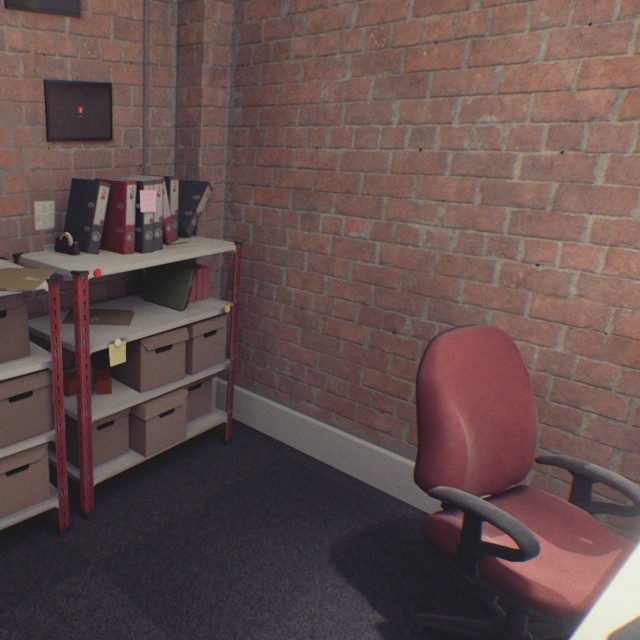}\\
  \smallskip
  \includegraphics[width=0.325\linewidth]{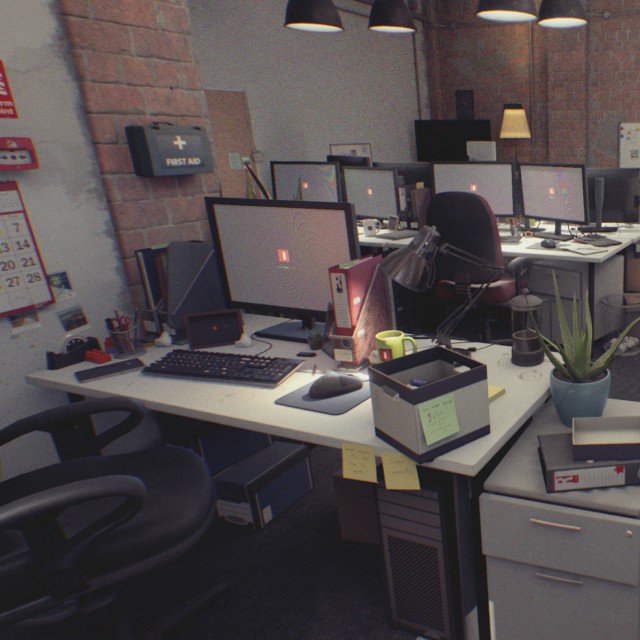}
  \includegraphics[width=0.325\linewidth]{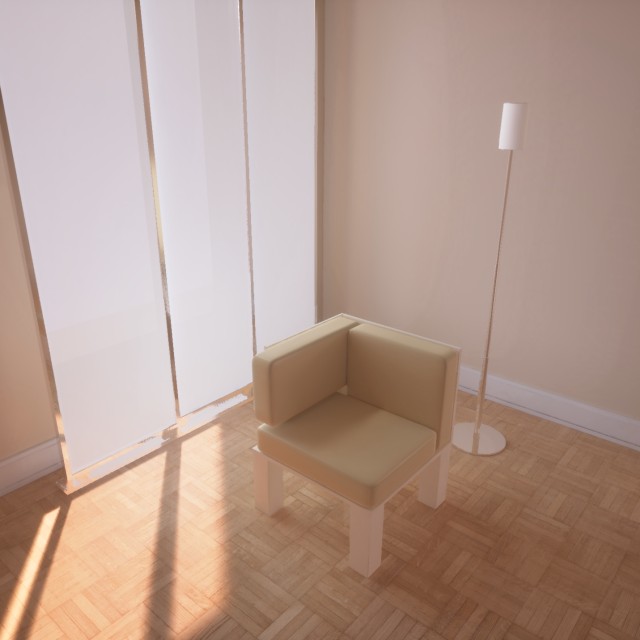}
  \includegraphics[width=0.325\linewidth]{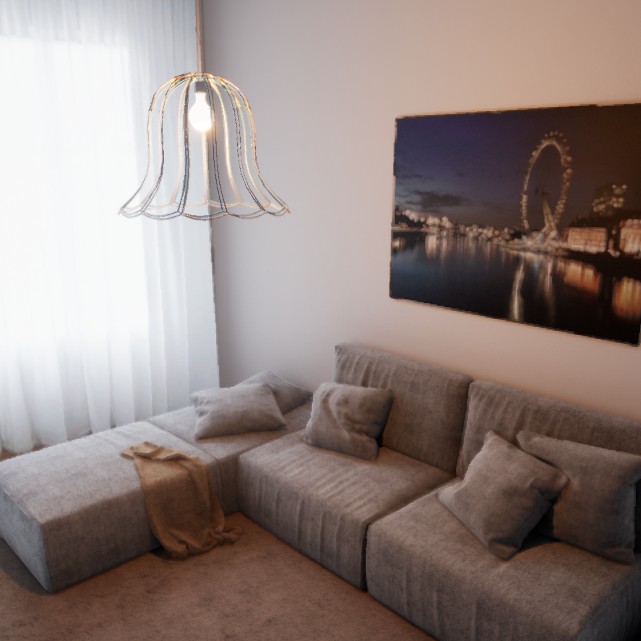}\\
  \smallskip
  \includegraphics[width=0.325\linewidth]{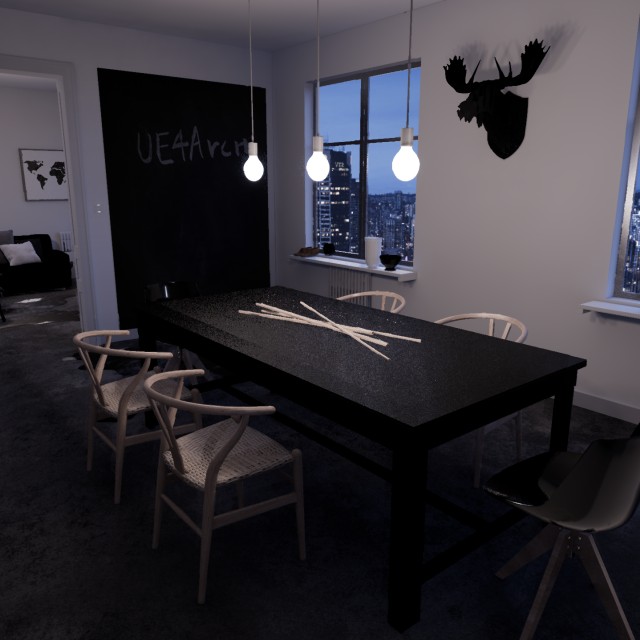}
  \includegraphics[width=0.325\linewidth]{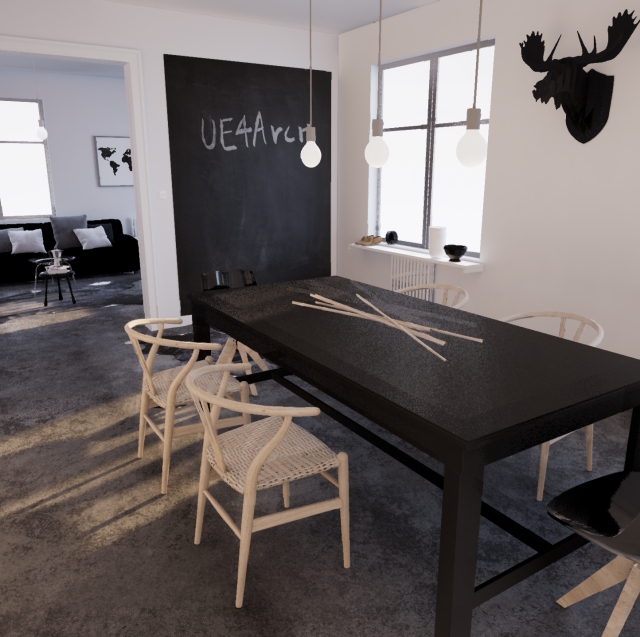}
  \includegraphics[width=0.325\linewidth]{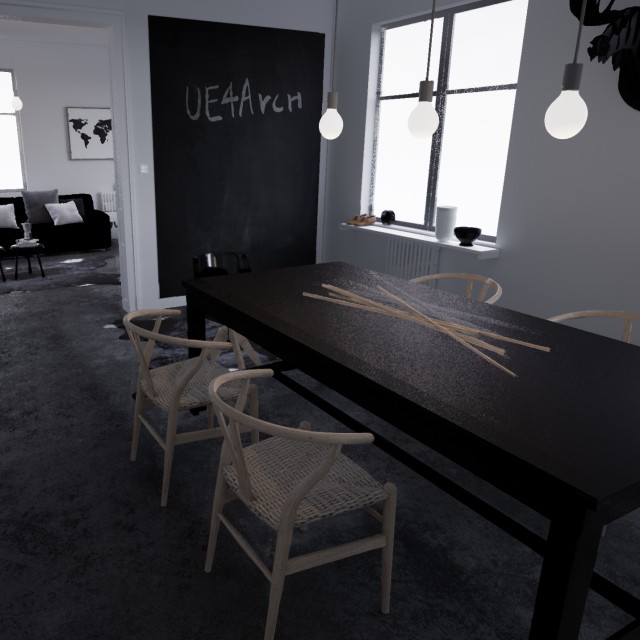}\\
  \caption{Snapshots of photorealistic scenes in the dataset.}
  \label{fig:scenes}
\end{figure}

\begin{itemize}
  \item 2D class mask @ $1920\times1080$ in 24-bits PNG.
  \item 2D/3D object instance oriented bounding boxes.
  \item 3D point cloud with RGB color.
  \item 3D instance/class mask.
\end{itemize}

This initial release of the dataset contains 32 detection classes and 39 semantic ones. These categories were selected from the most common and useful household goods in indoor environments for social robots. Note that semantic classes include structural elements such as walls that are not usually targets for detection that commonly focuses on relatively small and interactive objects. Table \ref{table:classes} shows both detection and semantic splits with their associated codes and colors.

All the tools, assets, and the dataset itself will be made available at \url{https://github.com/3dperceptionlab/therobotrix}.

\section{CONCLUSION}
\label{sec:conclusion}

In this work, we presented The RobotriX, an extremely realistic suite of data and tools designed to boost progress in indoor robotic vision tasks with deep learning by: (1) creating the largest and most realistic synthetic dataset to date; (2) seamlessly integrating realistic robots and scenes within virtual reality to easily generate plausible and useful sequences with interactions; (3) providing video sequences in multiple data modalities with perfect ground truth for solving and pushing forward the state of the art of a wide variety of problems, e.g., object detection, semantic segmentation, depth estimation, object tracking, object pose estimation, visual grasping, and many more. By releasing this dataset, our methodology, and the whole pipeline for generating data, we hope to satisfy the ever-growing need for data of deep learning approaches with easily generated and extremely realistic synthetic sequences which facilitate the deployment of those systems in real-world scenarios. 

As future works we plan on adding more complexity to the data and extend the range of problems that can benefit from it. For instance, we want to add non-rigid objects which can be simulated with \acl{UE4} physics such as elastic bodies, fluids, or clothes for the robots to interact with. We also want to automatically generate semantic descriptions for each frame to provide ground truth for captioning and question answering. In addition, we also want to add simulated force sensors on robotic hands to provide annotations for more sophisticated grasping tasks.

At last, we would like to remark that The RobotriX is intended to adapt to individual needs (so that anyone can generate custom data and ground truth for their problems) and change over time by adding new sequences thanks to its modular design and its open-source approach.

\section*{ACKNOWLEDGMENT}

This work has been funded by the Spanish Government TIN2016-76515-R grant for the COMBAHO project, supported with Feder funds. This work has also been supported by a Spanish national grant for PhD studies FPU15/04516 and by the University of Alicante project GRE16-19 and by the Valencian Government project GV/2018/022. Experiments were made possible by a generous hardware donation from NVIDIA.

\bibliographystyle{IEEEtran}
\bibliography{egbib}

\end{document}